\documentclass[runningheads]{llncs}
\usepackage[T1]{fontenc}
\usepackage{graphicx}
\usepackage{booktabs}
\usepackage{booktabs}
\usepackage{multirow}
\usepackage{tabularx}
\usepackage{makecell}
\usepackage{wrapfig}

\usepackage{booktabs}
\usepackage{tabularx}
\usepackage{makecell}
\usepackage[table]{xcolor}
\usepackage{siunitx}
\usepackage{subcaption}

\usepackage[table]{xcolor}   

\definecolor{propfill}{HTML}{E6F1FB}  
\definecolor{redgood}{HTML}{0F6E56} 
\usepackage{siunitx}
\usepackage{amssymb} 

\begin{document}
\title{Beyond Heuristics and Commercial Navigation: An AI-Augmented Decision-Support System for Ambulance Travel Time Estimation
}

\titlerunning{AI-Augmented Decision Support for Ambulance Travel Time Estimation}

\author{Muhammad Adil Abid \inst{1,2}\orcidID{0000-0002-0403-5353} \and
Johan Holmgren\inst{1,2}\orcidID{0000-0001-7773-9944} \and
Fabian Lorig \inst{1,2}\orcidID{0000-0002-8209-0921} \and
Jesper Petersson \inst{3,4}\orcidID{2222--3333-4444-5555}}
\authorrunning{M. A. Abid}
%
\institute{Department of Computer Science and Media Technology, Malmö University, 21119 Malmö, Sweden \and
Sustainable Digitalisation Research Centre, Malmö University, Sweden
 \and Department of Health Care Management, Region Skåne, 21428 Malmö, Sweden
 \and Department of Neurology, Lund University, 22242 Malmö, Sweden \\
\email{\{muhammad.adil-abid,johan.holmgren,fabian.lorig\}@mau.se}
\email{\{jesper.petersson\}@skane.se}}
\maketitle              
\begin{abstract}
Accurate ambulance travel time estimation is critical for emergency medical dispatch, yet existing decision-support practices often rely on static lookup tables or simplified assumptions regarding how ambulance travel time differs from ordinary passenger-car travel. For instance, they assume ambulances simply travel a fixed percentage (e.g., 5\%) faster than ordinary passenger cars. However, such assumptions are rarely validated using real ambulance operational data and may fail to capture the effects of trip priority, temporal context, and origin–destination. We address these limitations using historical ambulance trips from the Skåne region in Sweden. Our analysis reveals that ambulance travel times vary substantially by hour, day, and trip priority, and that commercial navigation systems fail to capture emergency-specific dynamics. We contribute a machine learning system to more accurately estimate ambulance travel time. To make the system interpretable, we explored SHapley Additive exPlanations and permutation importance techniques to ensure that our system captures operationally meaningful patterns.

\keywords{Ambulance travel time estimation \and Decision support systems  \and Emergency medical services  \and Explainable AI}
\end{abstract}

\section{Introduction}
\label{sec:introduction}
\footnotesize
In emergency medical services (EMS), ambulance travel time is a critical input for dispatch decisions as it affects how quickly prehospital care can reach a patient~\cite{intro-response-time-on-survival-rate-1}. Reliable travel time estimates can support ambulance selection, hospital routing, response planning, and resource allocation~\cite{intro-response-time-on-survival-rate-2}. However, ambulance travel time is affected by several contextual factors, including temporal conditions, trip urgency, and local road-network characteristics~\cite{intro-factors-influence-travel-time-1}. If such factors are ignored, dispatch systems may produce estimates that do not reflect the actual operating context, which can be especially problematic in time-sensitive cases~\cite{intro-ambulance-location-importnace}.

The EMS employs various systems to estimate ambulance travel time~\cite{Google-Maps-travel-time-estimation-1,saeid-five-percent-faster,adil-ambulance-travel time,sosalarm}. However, these systems remain limited because they often rely on (a) generic navigation estimates, designed for ordinary passenger-car traffic; (b) fixed-ratio speed assumptions, assuming that ambulances travel at a constant speed or fixed percentage faster than regular vehicles; or (c) predetermined lookup tables, assigning fixed travel times between locations. Such approaches may not fully capture ambulance specific movement patterns, trip priority, temporal variation, or local operational conditions. As a result, they can produce estimates that are insufficiently responsive to the dynamic and context-dependent nature of emergency medical services. This motivates the need for a data-driven system offering more accurate ambulance travel time estimates to support dispatch decisions.

We contribute a data-driven, context-aware travel time estimation system that learns from real-world ambulance operations and provides more accurate, interpretable estimates and illustrate its use in a scenario study of a regional EMS system in Sweden. Our system is based on machine learning, and it adapts its prediction according to the operational trip context, offering a promising alternative to static lookup tables, fixed-ratio heuristics, and generic navigation estimates. The proposed system computes the travel time for the ambulance in a given context, including variables that significantly improve the accuracy of the estimation. These variables are the origin and destination of each trip, trip priority, and departure time variables (\textit{e.g.,} hour of the day and day of the week). To identify these variables, we conducted an exploratory data analysis. Thereafter, we trained a supervised machine learning regressor on the identified variables. To make the system more operationally trustworthy, we explored two interpretability approaches, SHapley Additive exPlanations (SHAP)~\cite{intro-SHAP-xai} and permutation importance~\cite{intro-permutation-importance-xai}, to identify the contextual variables that most influence travel time estimates and to verify that the machine learning regressor has learned operationally meaningful patterns. The proposed system sets a new benchmark performance in estimating travel time, compared to a leading conventional system.

\section{Dataset and Exploratory Data Analysis}
\label{sec:dataset-eda}
\footnotesize

\subsection{Dataset}
\label{sec:dataset}
In this study, we use historical ambulance trip records from the Skåne region in southern Sweden covering $2020$ to $2021$. The data are provided by the regional emergency dispatch authority and contain operational information about completed ambulance trips. For the present study, we focused on variables that are available at dispatch time and can be interpreted operationally: origin area, destination area, trip priority, day of week, and hour of day. Trip priority is represented using three levels, where Priority $1$ denotes emergency trips, Priority $2$ denotes urgent trips, and Priority $3$ denotes routine or less-time-critical trips. The prediction target is ambulance travel time in minutes, measured as the elapsed time between the ambulance departure timestamp and the arrival timestamp. Cancelled trips and non-completed mission segments were filtered.

\subsection{Exploratory Data Analysis}
\label{sec:data-analysis}

The exploratory data analysis identifies periodic variation in ambulance travel time, which would invalidate static lookup tables or fixed-ratio assumptions.

\begin{figure}[t]
\centering
\begin{subfigure}[t]{0.45\linewidth}
    \centering
    \includegraphics[width=\linewidth]{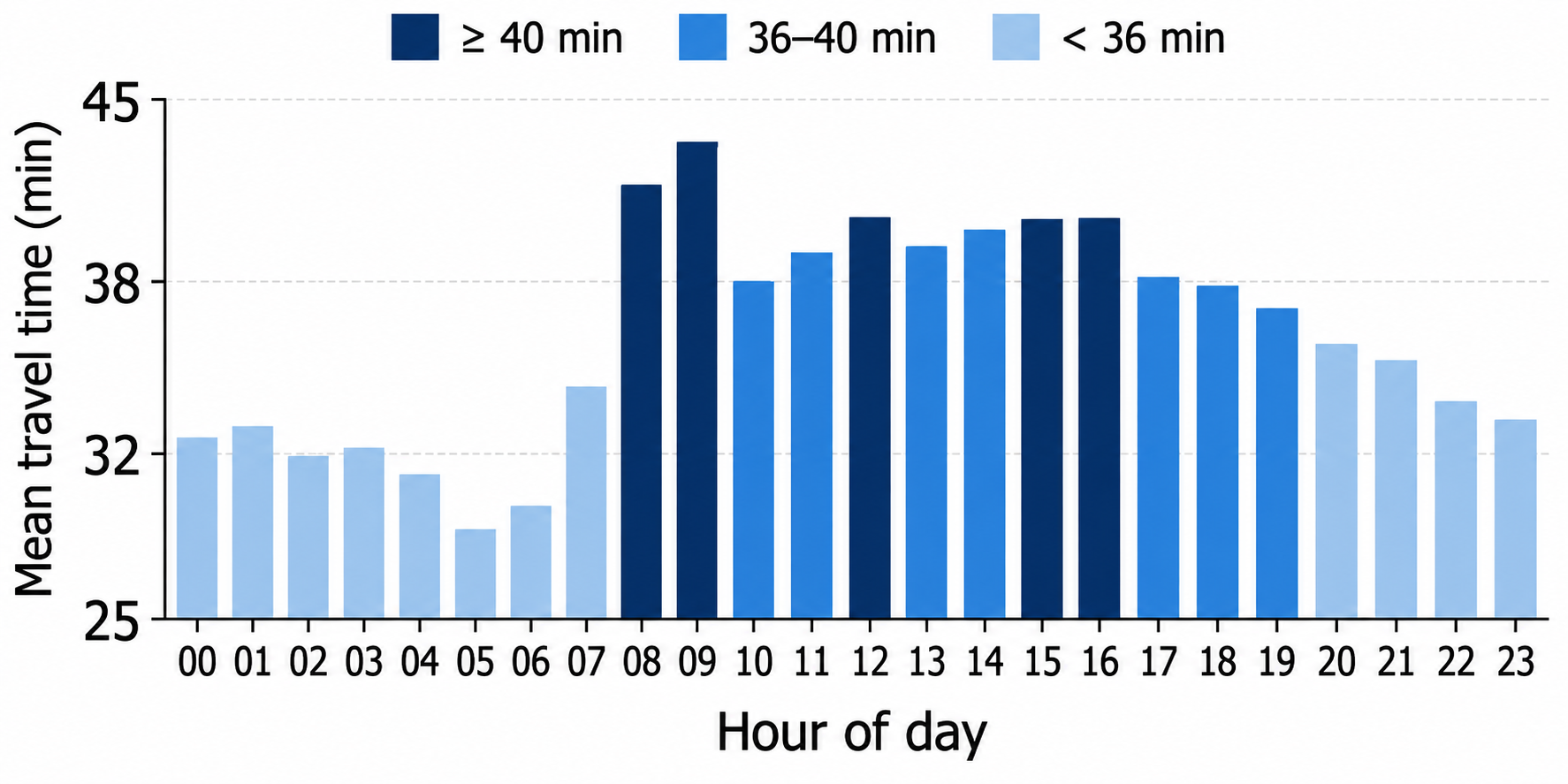}
    \caption{Hour of the day}
    \label{fig:eda-hour}
\end{subfigure}
\hfill
\begin{subfigure}[t]{0.23\linewidth}
    \centering
    \includegraphics[width=\linewidth]{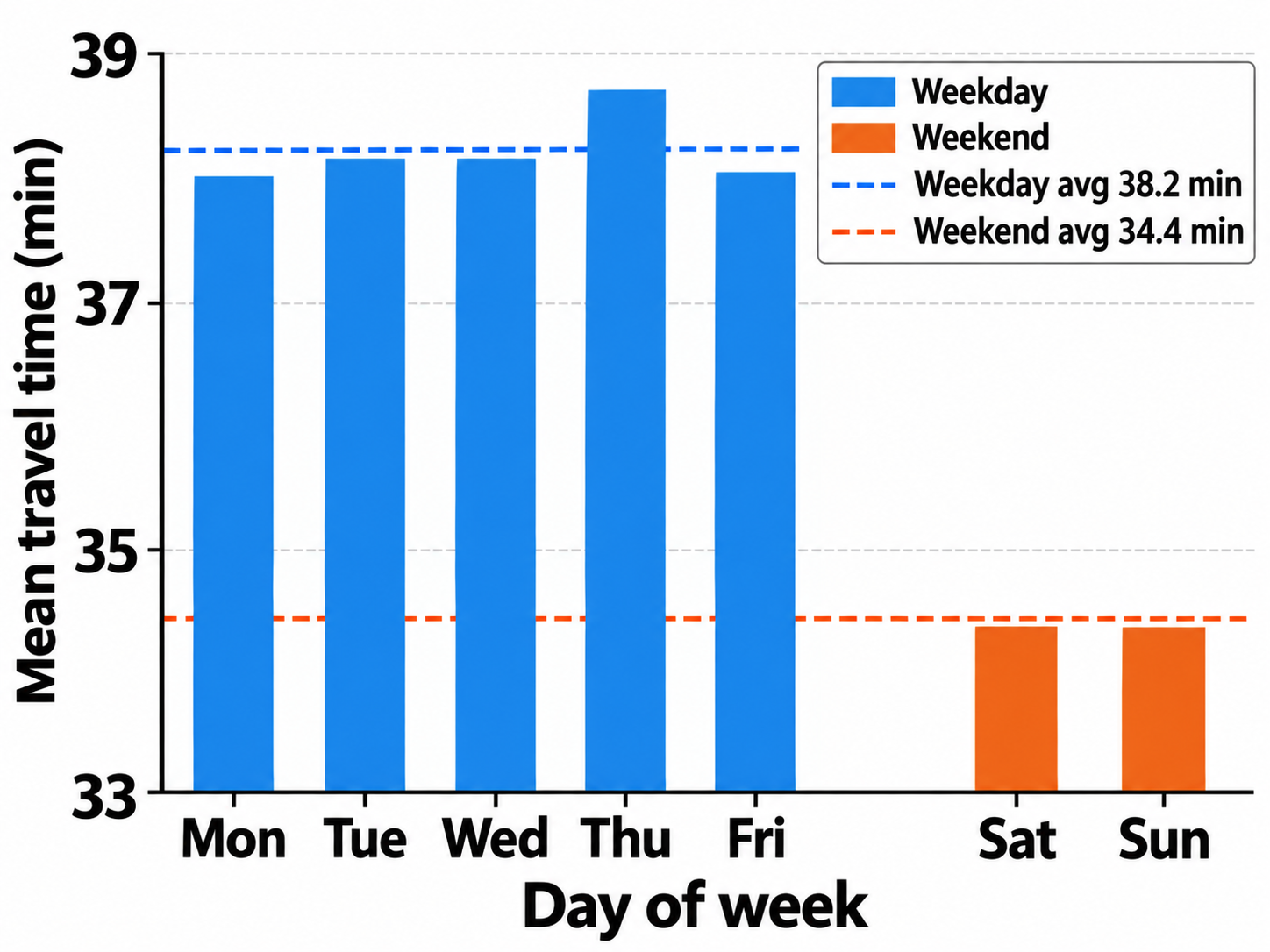}
    \caption{Day of the week}
    \label{fig:eda-day}
\end{subfigure}
\hfill
\begin{subfigure}[t]{0.2\linewidth}
    \centering
    \includegraphics[width=\linewidth]{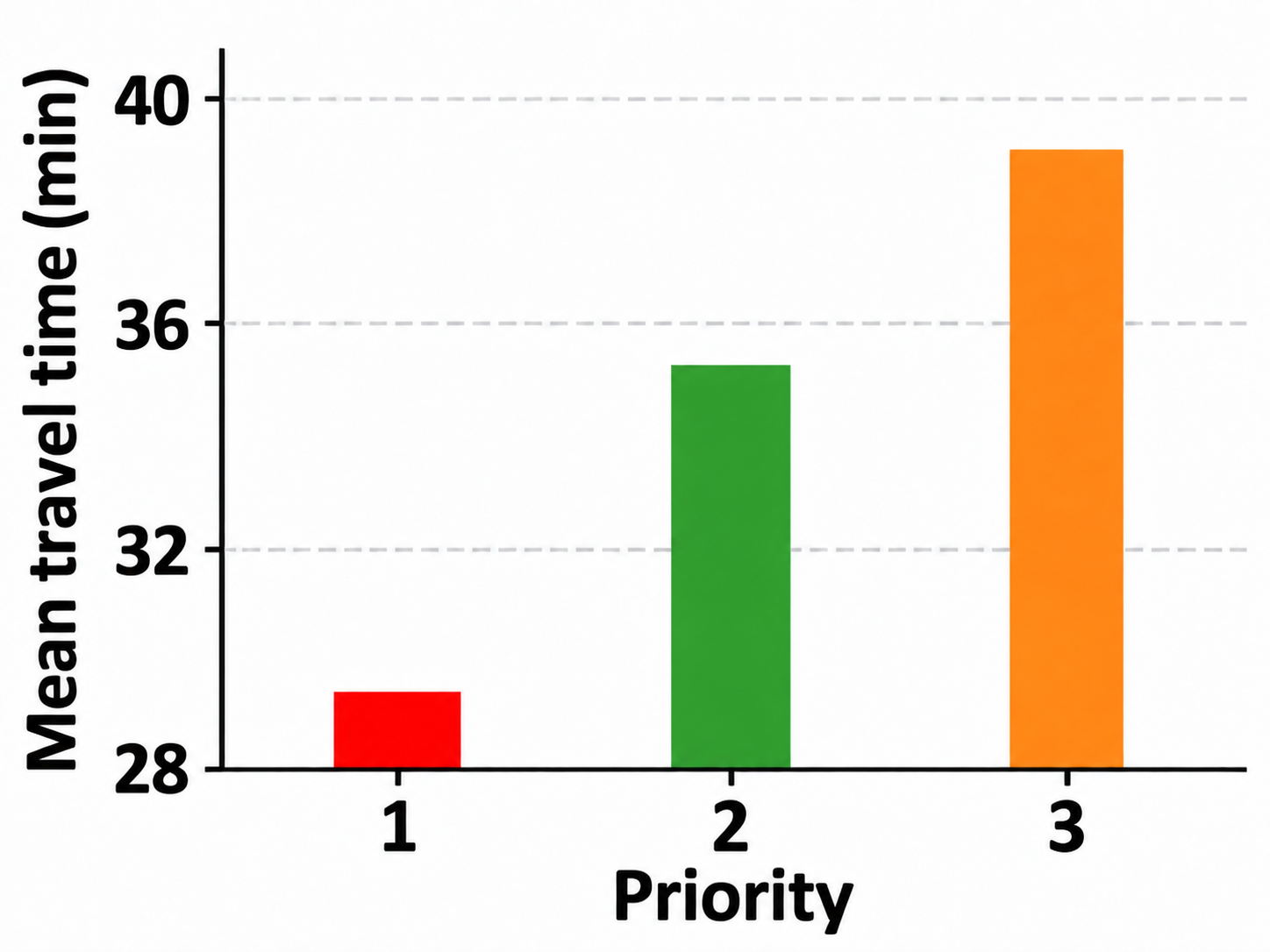}
    \caption{Priority level}
    \label{fig:eda-priority}
\end{subfigure}
\vspace{-1mm}
\caption{Mean ambulance travel time by (a) hour of the day, (b) day of the week, and (c) priority level.}
\label{fig:eda-overview}
\vspace{-2mm}
\end{figure}

\textbf{Hour-of-Day Patterns.}
\label{sec:eda-hour}
The analysis shows a substantial temporal range in the data (see Fig.~\ref{fig:eda-hour}): Ambulance travel times are lower during late-night and early-morning hours, rise sharply during the morning, and remain elevated through the daytime. The mean travel time is lowest at $05{:}00$ ($27.90$ min) and highest at $09{:}00$ ($43.35$ min), a $15.45$-minute difference.

\textbf{Day-of-Week Patterns.}
\label{sec:eda-day}
As shown in Fig.~\ref{fig:eda-day}, travel times are consistently higher on weekdays (mean 38.18 min) than on weekends (mean 34.38 min). Thursday is the day with the highest travel time (38.71 min), while Saturday and Sunday show similar values (34.38 and 34.37 min).

\textbf{Joint Hour-of-Day and Day-of-Week Patterns.}
\label{sec:eda-hour-day}
Joint hourly and daily patterns are more complex than a simple peak/off-peak split. On weekdays, mean travel times peak at $09{:}00$ and remain elevated until midday, with the highest value on Mondays ($48.5$ minutes). Weekend travel times are generally lower than on weekdays, and observed values range from $21.9$ to $48.5$ minutes. These patterns support retaining hour of day and day of week as separate features.

\textbf{Trip Priority.}
\label{sec:eda-priority}
Trip priority produces an even larger and operationally critical gap (see Fig.~\ref{fig:eda-priority}): Emergency trips (Priority $1$) complete in $29.3$ minutes on average, compared to $39.1$ minutes for routine trips (Priority $3$), nearly a $10$-minute difference, despite using the same road network, reflecting the influence of emergency driving practices such as lights-and-sirens operation and priority-aware driving behaviour. A decision-support system that is not conditional on priority level would systematically mispredict the highest-acuity, most time-sensitive trips.

Collectively, our analysis shows that ambulance travel time is jointly shaped by a fine-grained temporal context, trip priority, and route-specific factors. None of these dependencies can be represented by a fixed-ratio heuristic or a static lookup table, which by construction cannot vary with time, priority, or traffic conditions. This provides the empirical motivation for the data-driven method described in Section~\ref{sec:methodology}.

\section{AI-Augmented Decision Support for Ambulance Travel Time Estimation}
\label{sec:methodology}

\begin{figure}[t]
\centering
\includegraphics[width=1\linewidth]{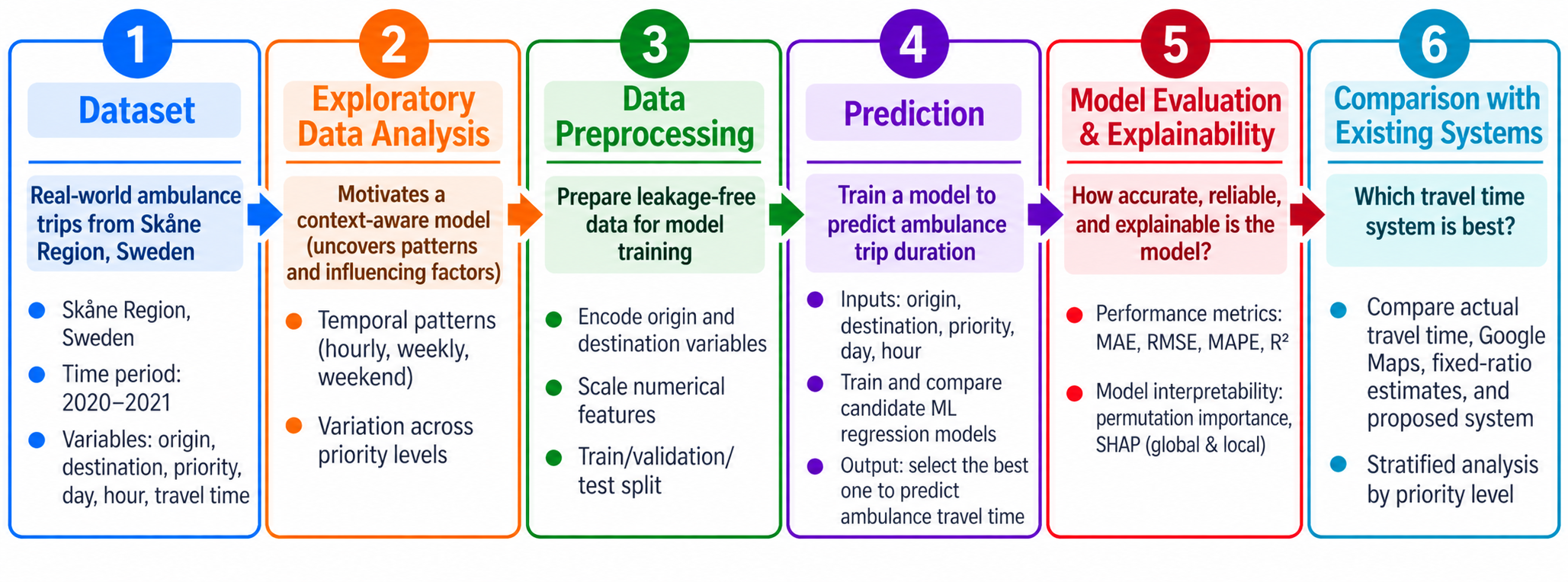}
\caption{A high-level overview of the proposed system’s methodological workflow.}
\label{fig:method-overview}
\end{figure}

Fig.~\ref{fig:method-overview} summarises the proposed system workflow followed in this study. Each step is detailed below.

\subsection{Data Preprocessing}
\label{sec:preprocessing}
The input data consist of five dispatch-available features: origin and destination area, trip priority, day of week, and hour of day. The two categorical spatial features, start-area and destination-area, are Swedish location names and are one-hot encoded. The remaining features are retained as numeric inputs, which are standardised using a standard scaler fitted only on the corresponding training partition to avoid data leakage. The resulting feature vector $\mathbf{x} \in \mathbb{R}^{d}$ is then passed to the machine learning regressor.

\subsection{Prediction}
\label{method:model-prediction}
We explored the following supervised machine learning models for ambulance travel time estimation: an artificial neural network (ANN), polynomial linear regression, CatBoost with native categorical handling, CatBoost with one-hot encoding, and random forest. These models represent different modelling assumptions: polynomial regression provides a simple statistical baseline; random forest and CatBoost capture non-linear tabular relationships; and the ANN is designed to learn complex interactions among spatial, temporal, and priority-related features.

\subsection{Explainability Methods}
\label{sec:xai-methods}

To move beyond black-box prediction and assess whether the proposed system learns operationally meaningful patterns, we applied two complementary interpretability techniques on the hold-out set: permutation feature importance and SHAP analysis.

\textbf{Permutation feature importance}. We estimated feature importance by measuring the increase in MAE when a feature is randomly permuted in the hold-out set, permuting all one-hot encoded columns of a categorical variable together as a block. To guard against unrealistic origin–destination structures from shuffling these independently, we also performed a joint origin–destination block permutation. Each permutation was repeated $N$ times, and we reported the mean MAE increase and standard deviation.

\textbf{SHAP}. We computed SHAP values using DeepExplainer with a background distribution of $N$ randomly sampled training instances, on the held-out test set. Because one-hot encoded spatial features can produce sparse explanations, SHAP values for origin and destination columns were aggregated to their parent feature level. We reported global feature importance via mean absolute SHAP values and analysed the combined contribution of hour of day and day of week to assess whether the system captures the temporal patterns identified in Section~\ref{sec:data-analysis}.

\subsection{Baseline Travel Time Estimation Methods}
\label{sec:benchmark-method}
To evaluate our proposed system against existing approaches, we used Google Maps estimates and fixed-ratio speed assumption estimates as baselines. We used the trained proposed system to estimate ambulance travel times for the hold-out test set and compared these estimates with the actual observed ambulance travel times, Google Maps estimates, and fixed-ratio speed assumption estimates. The comparison was stratified by trip priority to assess whether the relationship between commercial estimates and actual ambulance travel time is constant across priorities, as assumed by the fixed-ratio heuristic, or varies by operational context. Each baseline is detailed below.

\textbf{Google Maps Estimates}. We obtained commercial navigation estimates using the Google Routes API, requesting the route duration for each unique ``origin-destination-weekday-hour'' combination under driving mode with traffic-aware optimal routing, and converted the resulting durations from seconds to minutes. Since the original trips occurred in $2020$–$2021$ and the API does not support historical departure times, each request was instead assigned the next future occurrence of the trip's recorded weekday and hour (\textit{e.g.,} a Monday $09{:}00$ trip was queried for the next Monday at $09{:}00$); the benchmark therefore preserves the weekday-hour traffic pattern but should be interpreted as a matched temporal estimate rather than an exact historical reconstruction. Origins and destinations were represented by the area names available in the dataset, so the benchmark reflects area-level rather than door-to-door routing, consistent with the spatial granularity of the model inputs and the anonymisation of exact locations. Origin–destination pairs corresponding to identical or co-located facilities were assigned zero travel time and distance and excluded from the main interpretation, accounting for $806$ trips ($3.6$\% of the dataset).

\textbf{Fixed-ratio Speed Assumption Estimates.} For the second baseline, we followed the fixed-speed heuristic proposed by~\cite{saeid-five-percent-faster}, in which domain experts in stroke logistics assume that ambulances complete trips $5$\% faster than passenger cars. We adopted the same protocol and applied this adjustment to the Google Maps estimates, scaling each trip by $0.95$ to obtain the corresponding fixed-ratio speed travel time estimates for ambulances.

\section{Experiments and Results}
\label{sec:results}
\footnotesize

\subsection{Comparison of Candidate ML Models}
\label{sec:results-performance}
As shown in Table~\ref{tab:final-model-comparison} shows, the ANN achieved the best overall performance, with the lowest MAPE, MAE, and RMSE, as well as the highest $R^2$. Therefore, we selected the ANN as the prediction model for the proposed system, and we used its performance for the explainability analysis.

\subsection{Proposed System Comparison with Baseline Methods}
\label{sec:results-benchmark}

\begin{wrapfigure}{r}{0.6\textwidth}
    \centering
    \includegraphics[width=\linewidth]{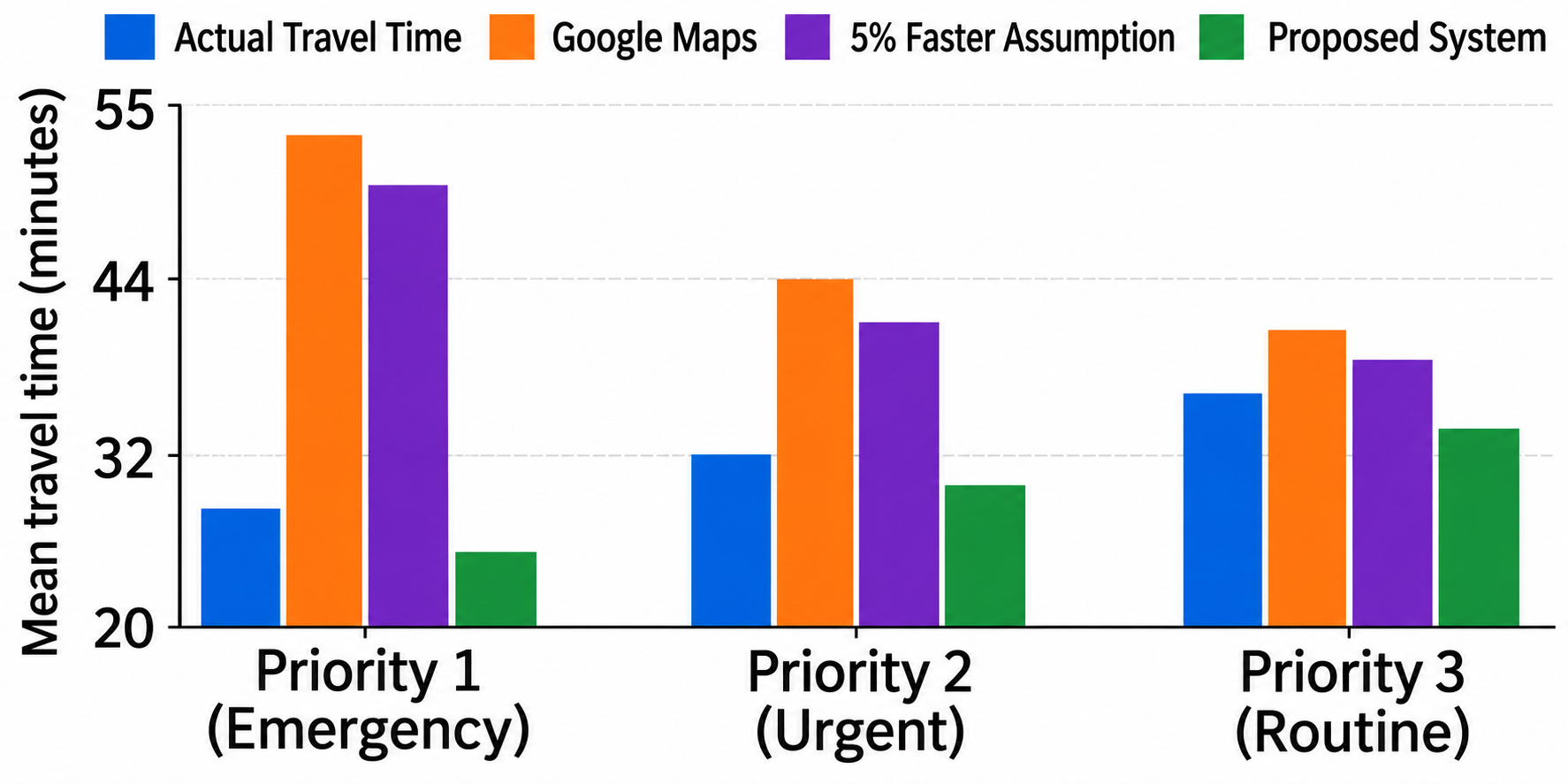}
    \caption{Benchmark comparison of mean ambulance travel time by priority level.}
    \label{fig:benchmark-google-ann}
\end{wrapfigure}

We compared the proposed system against Google Maps estimates and fixed-ratio speed assumptions from two perspectives: prediction error and mean travel time. Table~\ref{tab:mae_benchmark} reports the priority-stratified MAE, while Fig.~\ref{fig:benchmark-google-ann} compares actual ambulance travel time, Google Maps estimates, and the proposed system predictions across priorities.

The results show that both Google Maps and the $5$\% faster assumption overestimate ambulance travel time, particularly for higher-priority trips, whereas the proposed system remains closer to the observed travel times. The proposed system reduces MAE by $79.2$\%, $76.7$\%, and $52.7$\% compared with Google Maps for Priority 1, Priority $2$, and Priority $3$ trips, respectively, and it achieves similar reductions against the $5$\% faster baseline. This is operationally important because Priority~$1$ trips are precisely the cases where inaccurate travel time estimates can have the most serious consequences.

\begin{table}[t]
\centering
\scriptsize
\caption{Comparisons of the machine learning models}
\label{tab:final-model-comparison}
\begin{tabularx}{\columnwidth}{@{}l *{4}{>{\centering\arraybackslash}X}@{}}
\toprule
Model & MAPE (\%) $\downarrow$ & MAE (min) $\downarrow$ & RMSE (min) $\downarrow$ & $R^2$ $\uparrow$ \\
\midrule

Polynomial LR & $18.73$ & $5.24$ & $10.37$ & $0.65$ \\
CatBoost native & $13.20$ & $4.84$ & $10.72$ & $0.63$ \\
CatBoost one-hot & $13.90$ & $4.83$ & $10.50$ & $0.64$ \\
Random Forest & $19.76$ & $5.79$ & $11.18$ & $0.60$ \\
\textbf{ANN} & \textbf{12.51} & \textbf{4.72} & \textbf{10.15} & \textbf{0.67} \\
\bottomrule
\end{tabularx}
\end{table}

\definecolor{propfill}{HTML}{E6F1FB}   
\definecolor{hlgreen}{HTML}{DDE8D8}    

\begin{table}[t]
\centering
\scriptsize
\caption{Mean absolute error (MAE) of travel time estimates by priority.}
\label{tab:mae_benchmark}

\begin{tabularx}{\textwidth}{
    l
    *{2}{>{\centering\arraybackslash}X}
    >{\columncolor{propfill}\centering\arraybackslash}X
    @{\hspace{8pt}}
    *{2}{>{\columncolor{hlgreen}\centering\arraybackslash}X}
}
\toprule
& \multicolumn{3}{c}{MAE (min)}
& \multicolumn{2}{>{\columncolor{hlgreen}\centering\arraybackslash}p{0.30\textwidth}}
{\makecell{Proposed System\\ (Relative Error Reduction)}} \\
\cmidrule(lr){2-4} \cmidrule(lr){5-6}

Trip Priority 
& \makecell{Google\\ Maps}
& \makecell{$5$\% Faster\\ Assumption}
& \makecell{Proposed\\ System}
& \cellcolor{hlgreen}\makecell{vs. Google\\ Maps}
& \cellcolor{hlgreen}\makecell{vs. $5$\%\\ Assumption} \\
\midrule

P$1$ (Emergency) & $26.83$ & $24.53$ & \textbf{$5.58$} & $79.2$\% & $77.3$\% \\
P$2$ (Urgent)    & $19.50$ & $18.59$ & \textbf{$4.55$} & $76.7$\% & $75.5$\% \\
P$3$ (Routine)   & $10.11$ & $9.75$  & \textbf{$4.78$} & $52.7$\% & $51.0$\% \\

\bottomrule
\end{tabularx}
\end{table}

The results show that the relationship between Google Maps estimates and actual ambulance travel time is not a fixed offset but priority dependent. Although the $5$\% faster assumption provides slightly lower values than the Google Maps estimates, it still substantially overestimates ambulance travel time across all priorities, especially for Priority $1$ trips. Actual Priority 1 trips are completed in roughly $54$\% of the Google Maps estimate, Priority $2$ trips in about $70$\%, and Priority $3$ trips in about $87$\%. This priority-dependent pattern is far from a constant $5$\% adjustment and shows that no single fixed-ratio correction can accurately represent ambulance travel time across operational contexts. Therefore, a priority-aware estimation system is needed.

\subsection{Explainability Analysis}
\label{sec:results-interpretability}

To move beyond black-box prediction and assess whether the proposed system learns meaningful patterns, we applied permutation feature importance and SHAP analysis. These methods were applied to reveal which context features drive the proposed system’s predictions, strengthening the system’s transparency and supporting trust in its estimates.

\begin{figure}[t]
\centering
\begin{subfigure}[t]{0.35\linewidth}
    \centering
    \includegraphics[width=\linewidth]{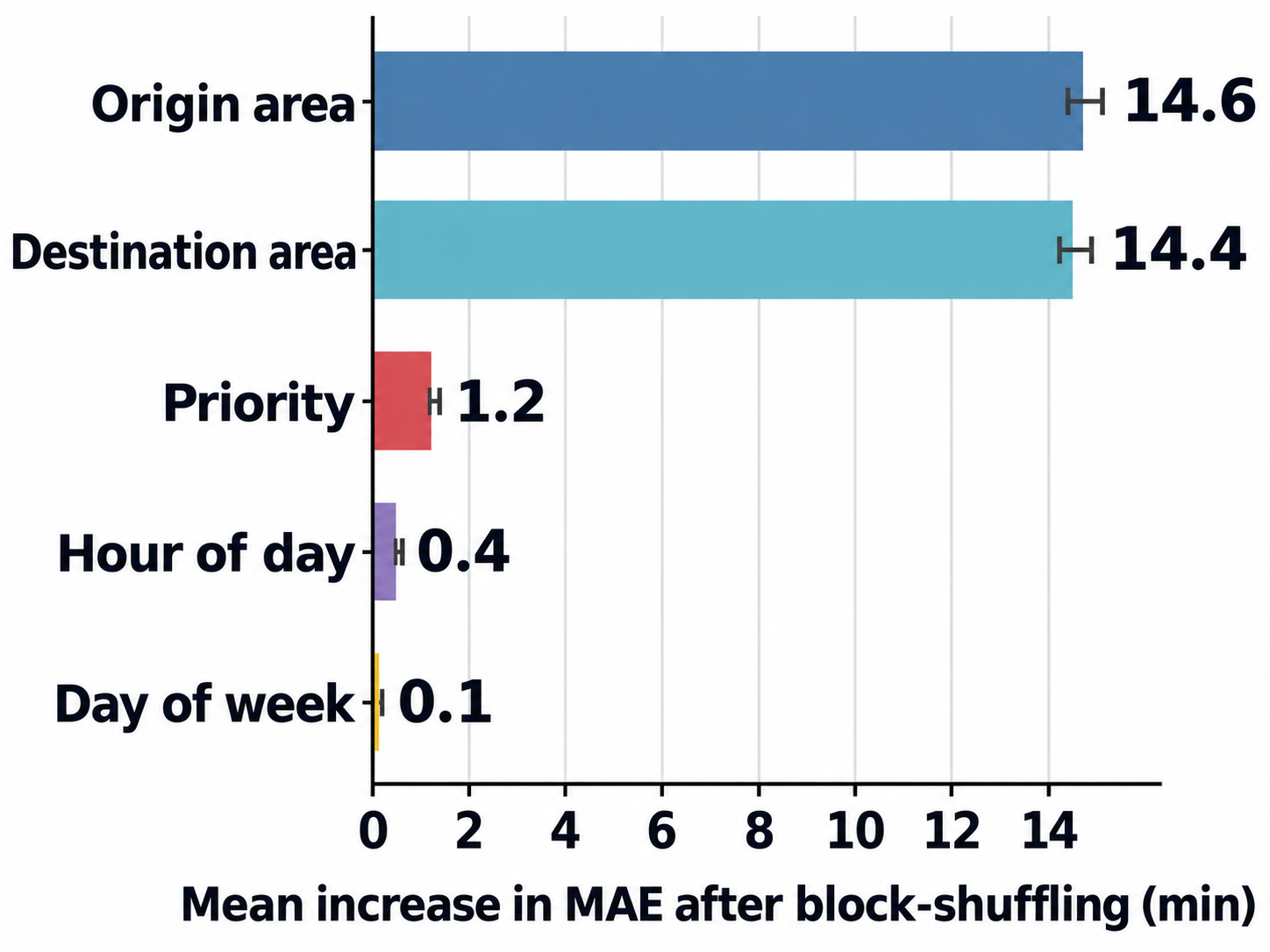}
    \caption{Permutation feature importance}
    \label{fig:permutation-importance}
\end{subfigure}
\hfill
\begin{subfigure}[t]{0.63\linewidth}
    \centering
    \includegraphics[width=\linewidth]{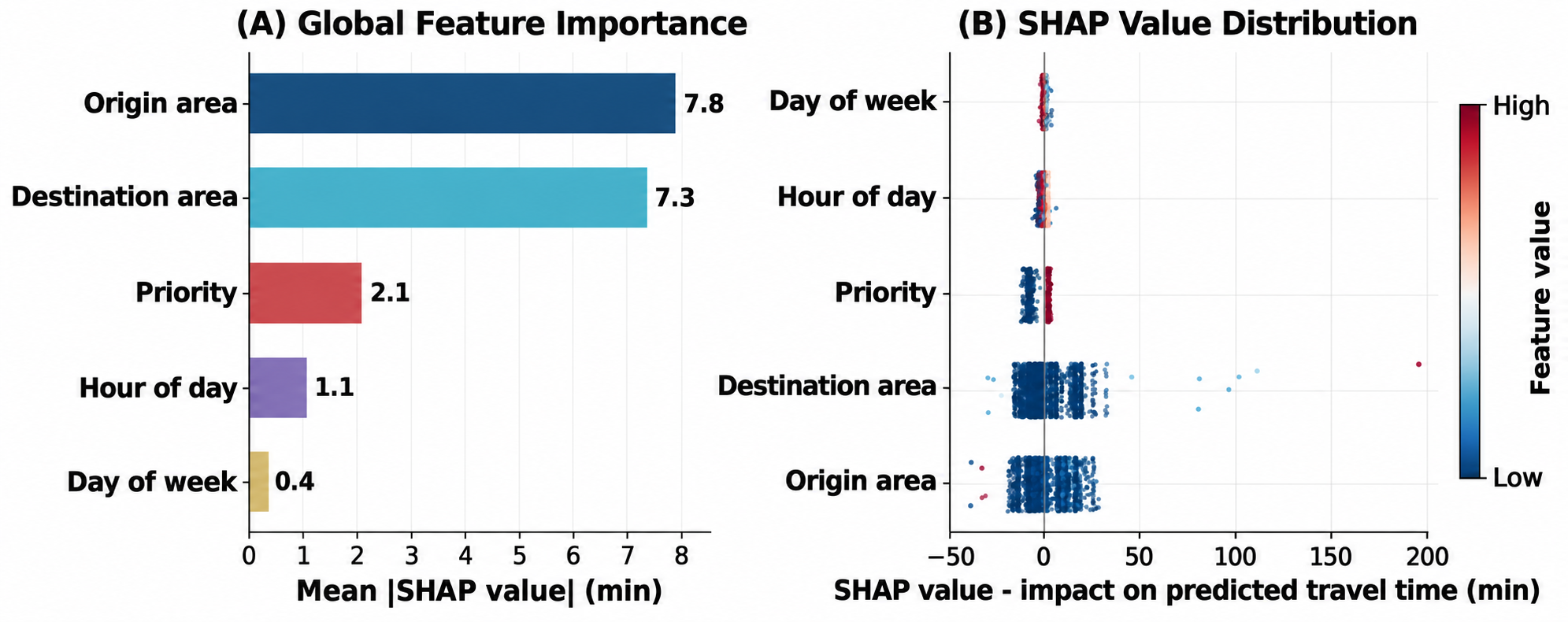}
    \caption{SHAP global explanations}
    \label{fig:shap-global}
\end{subfigure}
\vspace{-1mm}
\caption{Explainability system analysis: (a) permutation feature importance based on MAE increase after block shuffling and (b) SHAP global feature importance and value distribution.}
\label{fig:xai-analysis}
\vspace{-2mm}
\end{figure}

\textbf{Permutation Feature Importance.}
\label{sec:results-permutation}
The baseline MAE before permutation is $4.72$ minutes (see Fig.~\ref{fig:permutation-importance}). Origin area is the most influential feature, increasing the MAE by $14.6$ minutes when permuted, followed closely by destination area ($14.4$ minutes), priority ($1.2$ minutes), hour of day ($0.4$ minutes), and day of week ($0.1$ minutes). These results indicate that spatial context dominates ambulance travel time prediction, though the contributions of priority and temporal features show that travel time is not determined by route context alone. As a robustness check, a joint origin–destination pair permutation preserved valid travel routes, confirming that the high spatial importance reflects route-level structure rather than an artefact of independently disrupting origin and destination.

\textbf{Global SHAP Explanations.}
\label{sec:results-shap}
The global SHAP explanations (Fig.~\ref{fig:shap-global}) are consistent with the permutation analysis: origin area has the highest mean absolute SHAP value ($7.8$ minutes), followed by destination area ($7.3$), priority ($2.1$), hour of day ($1.1$), and day of week ($0.4$). These results further support the operational likelihood of the proposed system: origin and destination capture spatial trip structure, priority reflects differences among emergency, urgent, and routine trips, and hour of day aligns with the temporal variation observed in the exploratory analysis.

\section{Conclusions}
\label{sec:conclusion}
\footnotesize

This paper examined ambulance travel time estimation as a decision-support system in cooperative EMS dispatch systems. Using real-world dispatch data from the Sk\r{a}ne region in Sweden, we developed a data-driven, context-aware machine learning system. We showed that ambulance travel time varies substantially across spatial, temporal, and priority-related contexts, demonstrating the limitations of currently available systems for ambulance travel time estimations, which employ static lookup tables, fixed-ratio heuristics, and generic commercial navigation estimates. To support trustworthy decision support, we explored explainable AI components. The proposed system significantly outperforms systems that employ Google Maps estimates. We further showed that ambulance travel time cannot be represented by a single fixed-speed adjustment, as the gap between commercial estimates and actual ambulance travel varies systematically by urgency, with an implied speed advantage of roughly $46$\% for emergency trips versus $13$\% for routine ones. These findings support the need for ambulance-specific, interpretable, and priority-aware AI components for improving travel time estimation, transparency, and operational responsiveness in cooperative emergency dispatch systems. Accordingly, the paper offers a practical, low-latency component that can be integrated into EMS systems to reduce cognitive load and improve priority decisions.

\end{document}